%% file: main_paper.tex
\documentclass[5p]{elsarticle}

\usepackage{graphicx}
\usepackage{amsmath,amssymb}
\usepackage{algorithmic}
\usepackage{array}
\usepackage[caption=false,font=footnotesize,labelfont=sf,textfont=sf]{subfig}
\usepackage{fixltx2e}

\usepackage{comment}
\usepackage{epsfig}
\usepackage{changepage}
\usepackage{float}
\usepackage{booktabs}
\usepackage{xspace}

\usepackage{soul,xcolor}

\usepackage{lineno,hyperref}
\modulolinenumbers[5]

\journal{Pattern Recognition. DOI: \url{https://doi.org/10.1016/j.patcog.2021.108164}}
\license{Licensed \href{https://www.creativecommons.org/licenses/by-nc-nd/2.0/}{CC-BY-NC-ND 2.0}}

\begin{document}
\begin{frontmatter}
\title{Training Object Detectors from Few Weakly-Labeled and Many Unlabeled Images}

\author{Zhaohui Yang\fnref{pku}}
\author{Miaojing Shi\corref{corauthor}\fnref{miaojingshi}}
\author{Chao Xu\fnref{pku}}
\author{Vittorio Ferrari\fnref{vittorioferrari}}
\author{Yannis Avrithis\fnref{yannisavrithis}}

\cortext[corauthor]{Corresponding author. }
\fntext[pku]{Zhaohui Yang and Chao Xu are with Key Lab of Machine Perception, Dept. of Machine Intelligence, Peking University. Email: zhaohuiyang@pku.edu.cn, xuchao@cis.pku.edu.cn}
\fntext[miaojingshi]{Miaojing Shi is with King's College London. Email: miaojing.shi@kcl.ac.uk}
\fntext[vittorioferrari]{Vittorio Ferrari is with Google Research. Email: vittoferrari@gmail.com}
\fntext[yannisavrithis]{Yannis Avrithis is with Inria, Univ Rennes, CNRS, IRISA. Email: yannis@avrithis.net}

\begin{abstract}

Weakly-supervised object detection attempts to limit the amount of supervision by dispensing the need for bounding boxes, but still assumes image-level labels on the entire training set.
In this work, we study the problem of training an object detector from one or few images with image-level labels and a larger set of completely unlabeled images. This is an extreme case of semi-supervised learning where the labeled data are not enough to bootstrap the learning of a detector.
Our solution is to train a weakly-supervised student detector model
from image-level pseudo-labels generated on the unlabeled set by a teacher classifier model, bootstrapped by region-level similarities to
labeled images.
Building upon the recent representative weakly-supervised  pipeline PCL~\cite{tang2018pami},
our method can use
more unlabeled images to achieve
performance competitive or superior to many recent weakly-supervised detection solutions.
Code will be made available at \url{https://github.com/zhaohui-yang/NSOD}.
\end{abstract}


\end{frontmatter}

\input{abbrev}

\setstcolor{black}
\newcommand{\tea}[1][]{T_{#1}}
\newcommand{\stu}[1][]{U_{#1}}
\newcommand{\vsim}{\operatorname{sim}}
\newcommand{\vcls}{\sigma_{\mathrm{cls}}}
\newcommand{\vdet}{\sigma_{\mathrm{det}}}

\newcommand{\ours}{NSOD\xspace}
\newcommand{\oursg}{NSOD$_G$\xspace}
\newcommand{\oursx}{NSOD$_X$\xspace}

\newcommand{\yannis}[1]{{#1}}
\newcommand{\iavr}[1]{{#1}}
\newcommand{\zhaohui}[1]{{#1}}
\newcommand{\miaojing}[1]{{#1}}
\newcommand{\vitto}[1]{{#1}}

\input{introduction}

\input{relatedwork}

\input{method}

\input{experiment_short}

\input{discussion}

\input{acknowledgement}

\bibliography{references}

\end{document}

%% file: abbrev.tex

\newcommand{\head}[1]{{\smallskip\noindent\bf #1}}
\newcommand{\alert}[1]{{\color{red}{#1}}}
\newcommand{\eq}[1]{(\ref{eq:#1})\xspace}

\newcommand{\red}[1]{{\color{red}{#1}}}
\newcommand{\blue}[1]{{\color{blue}{#1}}}
\newcommand{\green}[1]{{\color{green}{#1}}}
\newcommand{\gray}[1]{{\color{gray}{#1}}}

\newcommand{\citeme}[1]{\red{[XX]}}
\newcommand{\refme}[1]{\red{(XX)}}


\newcommand{\tran}{^\top}
\newcommand{\mtran}{^{-\top}}
\newcommand{\zcol}{\mathbf{0}}
\newcommand{\zrow}{\zcol\tran}

\newcommand{\ind}{\mathbbm{1}}
\newcommand{\expect}{\mathbb{E}}
\newcommand{\nat}{\mathbb{N}}
\newcommand{\zahl}{\mathbb{Z}}
\newcommand{\real}{\mathbb{R}}
\newcommand{\proj}{\mathbb{P}}
\newcommand{\prob}{\mathbf{Pr}}

\newcommand{\mif}{\textrm{if }}
\newcommand{\minimize}{\textrm{minimize }}
\newcommand{\maximize}{\textrm{maximize }}

\newcommand{\id}{\operatorname{id}}
\newcommand{\const}{\operatorname{const}}
\newcommand{\sgn}{\operatorname{sgn}}
\newcommand{\var}{\operatorname{Var}}
\newcommand{\mean}{\operatorname{mean}}
\newcommand{\trace}{\operatorname{tr}}
\newcommand{\diag}{\operatorname{diag}}
\newcommand{\vect}{\operatorname{vec}}
\newcommand{\cov}{\operatorname{cov}}

\newcommand{\softmax}{\operatorname{softmax}}
\newcommand{\clip}{\operatorname{clip}}

\newcommand{\defn}{\mathrel{:=}}
\newcommand{\peq}{\mathrel{+\!=}}
\newcommand{\meq}{\mathrel{-\!=}}

\newcommand{\floor}[1]{\left\lfloor{#1}\right\rfloor}
\newcommand{\ceil}[1]{\left\lceil{#1}\right\rceil}
\newcommand{\inner}[1]{\left\langle{#1}\right\rangle}
\newcommand{\norm}[1]{\left\|{#1}\right\|}
\newcommand{\frob}[1]{\norm{#1}_F}
\newcommand{\card}[1]{\left|{#1}\right|\xspace}
\newcommand{\diff}{\mathrm{d}}
\newcommand{\der}[3][]{\frac{d^{#1}#2}{d#3^{#1}}}
\newcommand{\pder}[3][]{\frac{\partial^{#1}{#2}}{\partial{#3^{#1}}}}
\newcommand{\ipder}[3][]{\partial^{#1}{#2}/\partial{#3^{#1}}}
\newcommand{\dder}[3]{\frac{\partial^2{#1}}{\partial{#2}\partial{#3}}}

\newcommand{\wb}[1]{\overline{#1}}
\newcommand{\wt}[1]{\widetilde{#1}}

\def\xssp{\hspace{1pt}}
\def\ssp{\hspace{3pt}}
\def\msp{\hspace{5pt}}
\def\lsp{\hspace{12pt}}

\newcommand{\cA}{\mathcal{A}}
\newcommand{\cB}{\mathcal{B}}
\newcommand{\cC}{\mathcal{C}}
\newcommand{\cD}{\mathcal{D}}
\newcommand{\cE}{\mathcal{E}}
\newcommand{\cF}{\mathcal{F}}
\newcommand{\cG}{\mathcal{G}}
\newcommand{\cH}{\mathcal{H}}
\newcommand{\cI}{\mathcal{I}}
\newcommand{\cJ}{\mathcal{J}}
\newcommand{\cK}{\mathcal{K}}
\newcommand{\cL}{\mathcal{L}}
\newcommand{\cM}{\mathcal{M}}
\newcommand{\cN}{\mathcal{N}}
\newcommand{\cO}{\mathcal{O}}
\newcommand{\cP}{\mathcal{P}}
\newcommand{\cQ}{\mathcal{Q}}
\newcommand{\cR}{\mathcal{R}}
\newcommand{\cS}{\mathcal{S}}
\newcommand{\cT}{\mathcal{T}}
\newcommand{\cU}{\mathcal{U}}
\newcommand{\cV}{\mathcal{V}}
\newcommand{\cW}{\mathcal{W}}
\newcommand{\cX}{\mathcal{X}}
\newcommand{\cY}{\mathcal{Y}}
\newcommand{\cZ}{\mathcal{Z}}

\newcommand{\vA}{\mathbf{A}}
\newcommand{\vB}{\mathbf{B}}
\newcommand{\vC}{\mathbf{C}}
\newcommand{\vD}{\mathbf{D}}
\newcommand{\vE}{\mathbf{E}}
\newcommand{\vF}{\mathbf{F}}
\newcommand{\vG}{\mathbf{G}}
\newcommand{\vH}{\mathbf{H}}
\newcommand{\vI}{\mathbf{I}}
\newcommand{\vJ}{\mathbf{J}}
\newcommand{\vK}{\mathbf{K}}
\newcommand{\vL}{\mathbf{L}}
\newcommand{\vM}{\mathbf{M}}
\newcommand{\vN}{\mathbf{N}}
\newcommand{\vO}{\mathbf{O}}
\newcommand{\vP}{\mathbf{P}}
\newcommand{\vQ}{\mathbf{Q}}
\newcommand{\vR}{\mathbf{R}}
\newcommand{\vS}{\mathbf{S}}
\newcommand{\vT}{\mathbf{T}}
\newcommand{\vU}{\mathbf{U}}
\newcommand{\vV}{\mathbf{V}}
\newcommand{\vW}{\mathbf{W}}
\newcommand{\vX}{\mathbf{X}}
\newcommand{\vY}{\mathbf{Y}}
\newcommand{\vZ}{\mathbf{Z}}

\newcommand{\va}{\mathbf{a}}
\newcommand{\vb}{\mathbf{b}}
\newcommand{\vc}{\mathbf{c}}
\newcommand{\vd}{\mathbf{d}}
\newcommand{\ve}{\mathbf{e}}
\newcommand{\vf}{\mathbf{f}}
\newcommand{\vg}{\mathbf{g}}
\newcommand{\vh}{\mathbf{h}}
\newcommand{\vi}{\mathbf{i}}
\newcommand{\vj}{\mathbf{j}}
\newcommand{\vk}{\mathbf{k}}
\newcommand{\vl}{\mathbf{l}}
\newcommand{\vm}{\mathbf{m}}
\newcommand{\vn}{\mathbf{n}}
\newcommand{\vo}{\mathbf{o}}
\newcommand{\vp}{\mathbf{p}}
\newcommand{\vq}{\mathbf{q}}
\newcommand{\vr}{\mathbf{r}}
\newcommand{\Vs}{\mathbf{s}}
\newcommand{\vt}{\mathbf{t}}
\newcommand{\vu}{\mathbf{u}}
\newcommand{\vv}{\mathbf{v}}
\newcommand{\vw}{\mathbf{w}}
\newcommand{\vx}{\mathbf{x}}
\newcommand{\vy}{\mathbf{y}}
\newcommand{\vz}{\mathbf{z}}
\newcommand{\vone}{\mathbf{1}}
\newcommand{\vzero}{\mathbf{0}}

\newcommand{\valpha}{{\boldsymbol{\alpha}}}
\newcommand{\vbeta}{{\boldsymbol{\beta}}}
\newcommand{\vgamma}{{\boldsymbol{\gamma}}}
\newcommand{\vdelta}{{\boldsymbol{\delta}}}
\newcommand{\vepsilon}{{\boldsymbol{\epsilon}}}
\newcommand{\vzeta}{{\boldsymbol{\zeta}}}
\newcommand{\veta}{{\boldsymbol{\eta}}}
\newcommand{\vtheta}{{\boldsymbol{\theta}}}
\newcommand{\viota}{{\boldsymbol{\iota}}}
\newcommand{\vkappa}{{\boldsymbol{\kappa}}}
\newcommand{\vlambda}{{\boldsymbol{\lambda}}}
\newcommand{\vmu}{{\boldsymbol{\mu}}}
\newcommand{\vnu}{{\boldsymbol{\nu}}}
\newcommand{\vxi}{{\boldsymbol{\xi}}}
\newcommand{\vomikron}{{\boldsymbol{\omikron}}}
\newcommand{\vpi}{{\boldsymbol{\pi}}}
\newcommand{\vrho}{{\boldsymbol{\rho}}}
\newcommand{\vsigma}{{\boldsymbol{\sigma}}}
\newcommand{\vtau}{{\boldsymbol{\tau}}}
\newcommand{\vupsilon}{{\boldsymbol{\upsilon}}}
\newcommand{\vphi}{{\boldsymbol{\phi}}}
\newcommand{\vchi}{{\boldsymbol{\chi}}}
\newcommand{\vpsi}{{\boldsymbol{\psi}}}
\newcommand{\vomega}{{\boldsymbol{\omega}}}

\newcommand{\rLambda}{\mathrm{\Lambda}}
\newcommand{\rSigma}{\mathrm{\Sigma}}

\makeatletter
\newcommand*\bdot{\mathpalette\bdot@{.8}}
\newcommand*\bdot@[2]{\mathbin{\vcenter{\hbox{\scalebox{#2}{$\m@th#1\bullet$}}}}}
\makeatother

\makeatletter
\DeclareRobustCommand\onedot{\futurelet\@let@token\@onedot}
\def\@onedot{\ifx\@let@token.\else.\null\fi\xspace}
\def\eg{\emph{e.g}\onedot} \def\Eg{\emph{E.g}\onedot}
\def\ie{\emph{i.e}\onedot} \def\Ie{\emph{I.e}\onedot}
\def\cf{\emph{cf}\onedot} \def\Cf{\emph{C.f}\onedot}
\def\etc{\emph{etc}\onedot} \def\vs{\emph{vs}\onedot}
\def\wrt{w.r.t\onedot} \def\dof{d.o.f\onedot} \def\aka{a.k.a\onedot}
\def\etal{\emph{et al}\onedot}
\makeatother

%% file: introduction.tex
\section{Introduction}\label{sec:intro}

The objective of {visual object detection} is to place a tight bounding box on every instance of an object class. With the advent of deep learning, recent methods~\cite{frcnn,ren2015nips,chen2020robust,yuan2020gated} have significantly boosted the detection performance. Most are fully supervised, using a large amount of data with carefully annotated bounding boxes. However, annotating bounding boxes is expensive.

\iavr{To reduce}
the amount of supervision, the most common setting is \emph{weakly-supervised object detection} (WSOD)~\cite{wsddn,shi2016eccv,tang2017cvpr,cvpr20efficientwsod}. In this setting,
we are given a set of images known to contain instances of certain classes as specified by labels, but we do not know the object locations in the form of bounding boxes or otherwise. Many works~\cite{cinbis2014cvpr,shi2016eccv,shi2017arxiv,cao2017weakly}
formulate weakly supervised object detection as \emph{multiple instance learning} (MIL)~\cite{andrews2003nips}, which has been extended to be learnable end-to-end~\cite{wsddn,tang2017cvpr}.

There are \emph{mixed approaches} where a small number of images are annotated with bounding boxes and labels, and a large amount of images have only image-level labels~\cite{tang2016cvpr,hoffman2015cvpr,yan2017arxiv}.
This is often referred as a \emph{semi-supervised} setting~\cite{tang2016cvpr,yan2017arxiv,huang2021behavior}, but there is no consensus.

\emph{Semi-supervised learning}~\cite{CSZ06} refers to using a small amount of labeled data and a large amount of unlabeled data. It is traditionally studied for classification~\cite{WeRC08,Lee13,RBH+15}, with one class label per image and no bounding boxes. In object detection, this would normally translate to a small number of images having labels {and} bounding boxes, and a large number of images having {no annotation at all}. This problem has been studied for the case where the fully annotated data (with bounding boxes) are enough to train a detector in the first place~\cite{1802.06964,RDG+18}, resulting in two-stage learning. But what if these data are very scarce?

\begin{figure}[t]
	\centering
	\includegraphics[width=1\columnwidth]{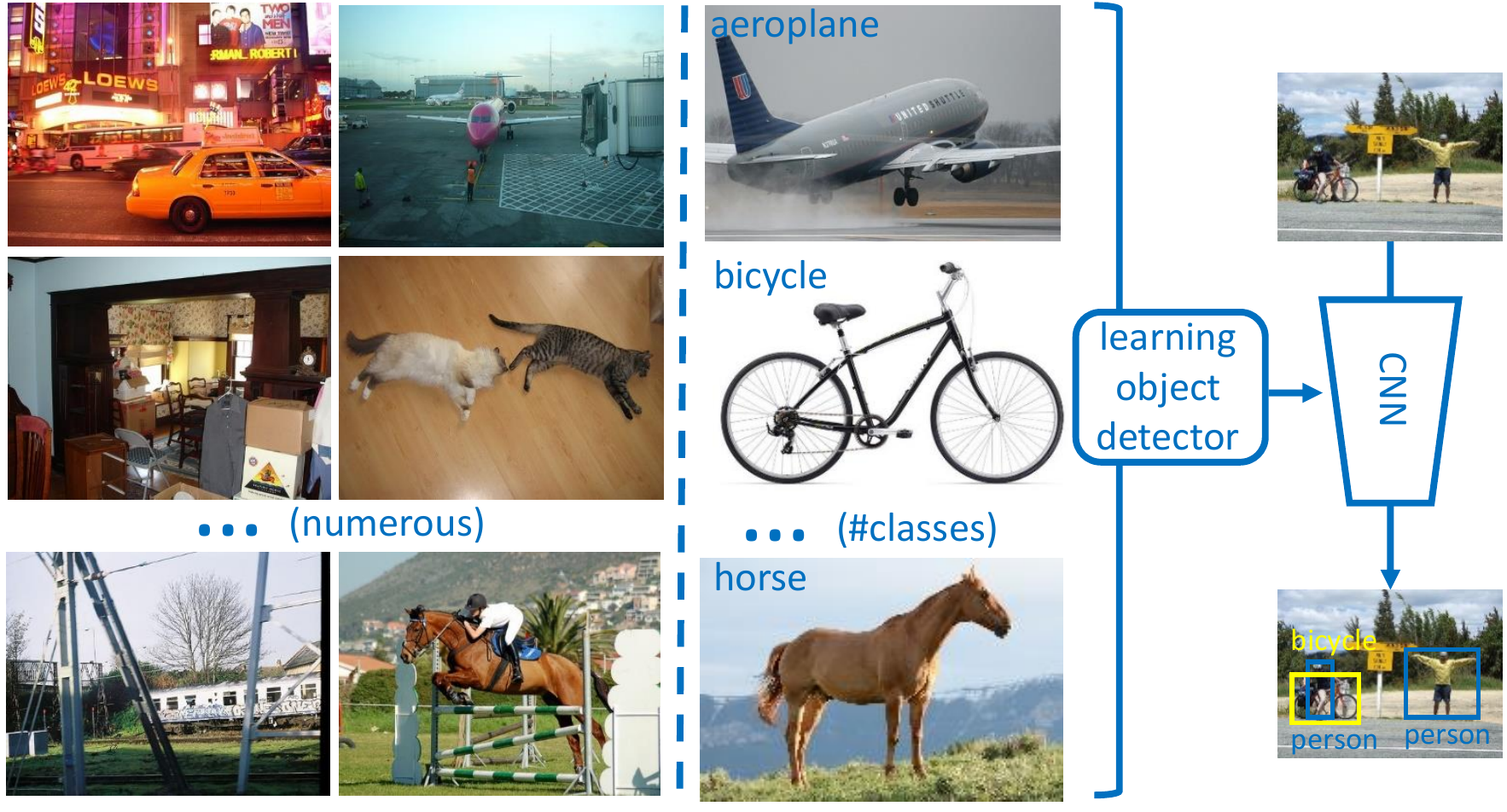}
	\caption{We learn an object detector from a set of completely unlabeled images and one or few images per class with image-level label per image and no other information.
	}
	\label{Fig:moti}
\end{figure}

\miaojing{In this work, we study object detection in the challenging setting where only one or few images per class are given with only image-level class label per image, and a large amount of images with no annotation at all.
We use no bounding boxes or other information. This setting is illustrated in Fig.~\ref{Fig:moti}.
Some initial exploration can be found in~\cite{ShHX15,marvaniya2012drawing,modolo2017pami} before deep learning.
The few weakly-labeled images can be obtained via either labeling
images from an unlabeled collection~\cite{ShHX15,marvaniya2012drawing} or using
the top-ranking images from \emph{web image search} with the class name as the query~\cite{modolo2017pami}. Both paradigms are studied in our work.
The latter is preferable as it requires no human effort.
}

Our \miaojing{deep learning} solution is called \emph{nano-supervised object detection} (NSOD).
It begins by computing region-level class scores based on the similarity
between the unlabeled images and
{the few weakly-labeled images}, which we then pool into image-level class probabilities.
{This yields image-level \emph{pseudo-labels} on the entire unlabeled set, which we use to train a \emph{teacher} model on a classification task.}
Then, by predicting new image-level \iavr{multi-class} pseudo-labels on the unlabeled set, we train a \emph{student} model on a detection task, using a weakly-supervised object detection pipeline.

\head{Contributions.}
\miaojing{We study the very challenging
problem} of training
an object detector from few
images with only image-level labels and many images with no annotation at all. We introduce a new method for this problem that is \emph{simple}, \emph{efficient} (cost comparable to standard WSOD), and \emph{modular} (can build on any WSOD pipeline). By using the recent pipeline of PCL~\cite{tang2018pami} and more unlabeled images, we achieve performance competitive or superior to many recent WSOD solutions.
{On PASCAL VOC 2007 test set for instance, using 20 web images per class,
we get a detection mAP of 42, compared to 43.5 of PCL, which is using image-level labels on the entire training set.
}

%% file: relatedwork.tex
\section{Related Work}
\label{sec:related}

\subsection{Weakly supervised object detection (WSOD).}
In this setting, all training images have image-level class labels. A classic approach is
\emph{multiple instance learning} (MIL)~\cite{andrews2003nips}, considering each training image as a ``bag'' and iteratively selecting {high-scoring object proposals}
from each bag, treating them as ground truth to learn an object detector.

Bilen and Vedaldi~\cite{wsddn} introduce \emph{weakly-supervised deep detection network} (WSDDN), which pools region-level sco\-res into image-level class probabilities and enables end-to-end learning from image-level labels. Concurrently, Tang~\etal~\cite{tang2017pr} introduce a deep convolutional neural network by integrating traditional multiple instance learning (MIL) into it for end-to-end training. Furthermore, Tang~\etal~\cite{tang2017cvpr} extend WSDDN to multiple instance detection network with an \emph{online instance classifier refinement} (OICR) sche\-me and introduce a weakly-supervised {region proposal network} as a plugin~\cite{tang2018eccv}. In \emph{proposal cluster learning} (PCL)~\cite{tang2018pami}, pre-clustering of object proposals followed by OICR accelerates learning and boosts performance.
In~\cite{zhang2018weakly}, a pseudo ground truth mining algorithm is also introduced to improve OICR.
Recently, Zeng~\etal~\cite{zeng2019wsod2} propose a novel WSOD framework with objectness distillation by jointly considering bottom-up and top-down objectness from low-level measurement and CNN confidences with an adaptive linear combination.
Ren~\etal~\cite{cvpr20efficientwsod} employ an instance-aware self-training strategy for WSOD with Concrete DropBlock.
Zhang~\etal~\cite{zhang2021cadn} extract the category-aware spatial information from a classification network to both classify and localize objects using image-level annotation. Liu~\etal~\cite{liu2021weakly} leverage a graph neural network into WSOD to discover semantic label co-occurrence.

Besides improvements in the network architecture, there are also attempts to incorporate additional cues into WSOD that are still weaker than bounding boxes, \eg object \emph{size}~\cite{shi2016eccv} and \emph{count}~\cite{gao2018eccv}.
It is also common to use extra data to \emph{transfer knowledge} from a source domain and help localize objects in the target domain~\cite{shi2017iccv,uijlings2018revisiting}.
Large-scale weakly-labelled \emph{web images}~\cite{guo2018eccv,tao2018tmm} and \emph{videos}~\cite{singh2019cvpr,liang2015cvpr}
with noisy labels are also common as extra data.

Our problem is different from WSOD in that the few labeled images have no bounding boxes and the bulk of the training set is completely unlabeled.
We build our work on PCL~\cite{tang2018pami} but train it with image-level pseudo-labels.

\subsection{Semi-supervised learning.}
There are several works that assume a few images are annotated with object bounding boxes and the rest still have image-level labels as in WSOD~\cite{tang2016cvpr,hoffman2015cvpr,yan2017arxiv}. These are often called \emph{semi-supervised}~\cite{tang2016cvpr,yan2017arxiv, sheikhpour2017a, cevikalp2020semi,gao2019iccv}. However, semi-supervised may also refer to the situation where some images are labeled (at image-level or with bounding boxes) and the rest have {no annotation at all}~\cite{1802.06964,RDG+18}. This situation is consistent with the standard definition of \emph{semi-supervised learning}~\cite{CSZ06}.
{Despite advances in deep semi-supervised learning~\cite{hoffer2016arxiv,LA17,TV17}, most work focuses on classification tasks.
In \emph{pseudo-label}~\cite{Lee13} for instance, classifier predictions on unlabeled data are used as labels along with true labels on labeled data.}
Few exceptions focusing on object detection~\cite{1802.06964,RDG+18,tang2021wacv,gao2019iccv} still assume enough labeled images to learn a detector in the first place, which is not the case in our work.

Tang~\etal~\cite{tang2021wacv} assume part of the training set is strong\-ly labelled with bounding boxes and the other part is unlabelled. They also call this setting semi-supervised.
They experiment on large labeled and unlabeled sets (118K and 123K respectively in COCO): the labeled data is trained in a standard Faster R-CNN,  while for the unlabeled data, a self-supervised proposal learning module and a consistency-based proposal learning module are introduced. Gao~\etal~\cite{gao2019iccv} assume a few seed training images are annotated with bounding boxes and the rest are weakly labeled with image-level annotations. This is again called semi-super\-vised. The seed samples are trained with Faster R-CNN while an iterative training-mining pipeline is introduced to mine bounding boxes from the weakly-labelled set for joint training.

The supervision settings of Tang~\etal~\cite{tang2021wacv} and Gao~\etal~\cite{gao2019iccv} are different, but in both cases, the labelled data are enough to bootstrap a standard Faster R-CNN. This is not the case in our work. Dong \etal~\cite{dong2018pami} use few images with object bounding boxes and class labels along with many unlabeled images. However,
this method
relies on several models (\ie Fast R-CNN, R-FCN, SPL) and iterative training, which is computationally expensive.

Our problem can be considered as an extreme case of semi-supervised object detection:  the labeled images are very few and with only image-level labels, which is too little to learn a good detector like Faster R-CNN. We thus introduce a \emph{nano-supervised} solution with teacher-student distillation. \iavr{Shi \etal~\cite{ShHX15} also use a mixture of few weakly-labeled images and unlabeled images for object detection. Their method involves hand-crafted features and iterative message passing, which would not be straightforward or efficient to extend to a deep learning framework.

It should be noted that Gao~\etal~\cite{gao2019iccv} also employ a teacher model in their pipeline, but the teacher is an object detector pre-trained on a large amount of fully-labelled images on source classes. This provides additional help against the \emph{noisy labels} in the bounding box mining process. By contrast, our teacher comes from a model pre-trained on ILSVRC classification, which is the most easily and widely accessed model for the majority of computer vision tasks. Thanks to our careful design of knowledge distillation, our approach  also turns out to be effective and robust to noisy labels.}

\begin{figure*}[t]
	\centering
	\includegraphics[width=0.85\textwidth]{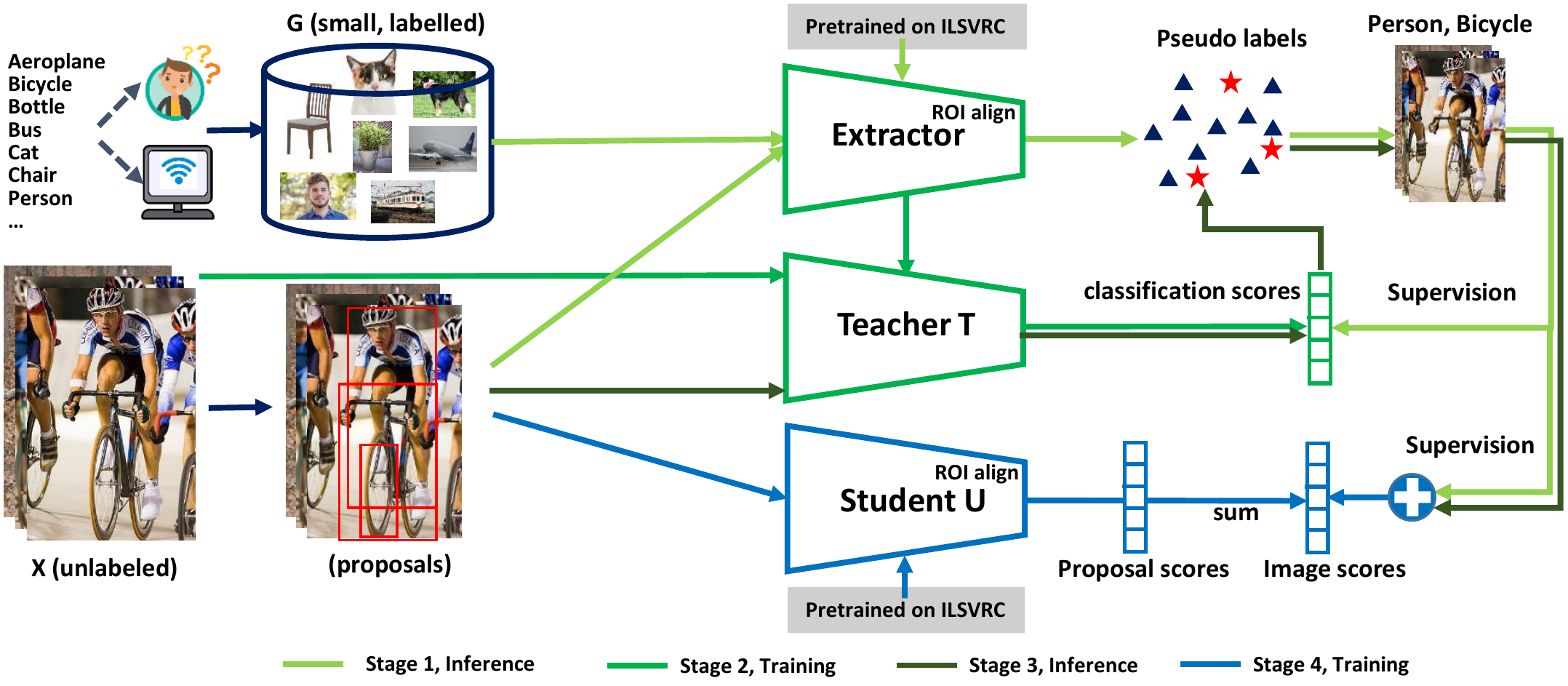}
	\caption{\small Overview of our \emph{nano-supervised object detection} (NSOD) framework. We are given a support set $G$ and a large unlabeled set $X$. {$G$ contains one or few weakly-labeled images per class, obtained from the web or randomly labeled from $X$}.
	{Using the images in $G$ and a feature extractor pre-trained on classification, we infer image-level class probabilities of images in $X$}
	(stage 1). We then extract pseudo-labels on $X$ and
	{train a \emph{teacher} network}
	$T$ on a $C$-way classification task (stage 2).
	$T$ is used to classify each proposal of images in $X$, resulting in new image-level class probabilities (stage 3). We average these with the ones obtained in stage 1, based on $G$. Finally, we extract multi-class pseudo-labels on $X$ and train a \emph{student} network $\stu$ {on weakly-supervised detection} by PCL~\cite{tang2018pami} (stage 4).}
	\label{Fig:intro}
\end{figure*}

\subsection{\iavr{Curated data.}}
\iavr{
Investigation of unsupervised settings relies on removing the labels from labeled datasets by default. This is the case \eg for \emph{object discovery}~\cite{TLBB10,sivic2005iccv}, 
\emph{semi-supervised classification}~\cite{berthelot2019mixmatch} and crowd counting~\cite{zhao2020eccv,liu2020acmmm} until today. Such datasets are \emph{curated}, \ie, still depict the same classes and are more or less balanced. Working with unknown classes is a different problem of \emph{open-set recognition}~\cite{bendale2016towards}. At very large scale, keeping the top-ranking examples according to predicted class scores may be enough to address this problem~\cite{yalniz2019billion}. We experiment on both curated and unlabeled data \emph{in the wild} to show the robustness of our method.
}

%% file: method.tex
\section{Method}
\label{sec:method}

\subsection{Preliminaries}

\head{Problem.}
We are given a \emph{support set} $G$ \iavr{containing $k$ images per class}, each associated with an image-level label over $C$ classes. 
We are \iavr{also} given \iavr{an \emph{unlabeled}} set of images $X$, 
where each image \iavr{depicts} one or more 
\iavr{instances}
of \iavr{the} $C$ classes, along with background clutter.
\iavr{In a harder setting, images in $X$ may depict zero or more instances of the $C$ classes, along with instances of unknown classes or background clutter.}
There is no bounding box or any other information in either $G$ or $X$.
Using these data and a {feature extractor $\phi$ pre-trained on classification},
the problem is to learn a detector
to recognize instances of the $C$ classes and localize them with bounding boxes in new images.

\head{Motivation.}
This problem relates to both weakly-super\-vised detection and semi-supervised classification.
Similar to the former, we study multiple instance learning but without image-level labels in the unlabeled set. 
\iavr{Unlike} 
the latter, at least in its common setting where thousands of examples are used~\cite{RBH+15,LA17}, 
$G$ is too small to bootstrap the learning of a good classifier or detector:
$k$ can be as few as one example per class. For this reason, we
propagate labels from $G$ to $X$ to initiate training.

\head{Method overview.}
{As shown in Fig.~\ref{Fig:intro}, we}
\iavr{begin}
by collecting \iavr{the support set} $G$
(Sec.~\ref{Sec:setup}).
We extract
{object proposals}~\cite{edgeboxes} from images in $X$ and compare region-level features 
\iavr{obtained by}
a feature extractor $\phi$
against global features 
on
$G$. We estimate class probabilities on $X$ by propagating these similarities to image level (stage 1,
\iavr{Sec.~\ref{Sec:webassignment}}).
We infer pseudo-labels on $X$ and train a \emph{teacher} network $\tea$ inherited from $\phi$ on a $C$-way classification task (stage 2, Sec.~\ref{Sec:teacherstudent}).
\iavr{We use $T$} 
to classify regions in images of $X$, resulting in new image-level class probabilities (stage 3), which we average with the ones 
of
stage 1.
Finally, we \iavr{infer} multi-class pseudo-labels on $X$ and train a \emph{student} network $\stu$ {on a WSOD task} by PCL~\cite{tang2018pami} (stage 4).

\head{\iavr{Collecting the} support set $G$.}
\miaojing{The support set can be obtained either by
{random selection
from some existing dataset or by web image search.
The latter is preferable as 
we would like 
images to be clean, \eg depicting only one class per image.%
} 
We experiment with both options.
}

\subsection{Inferring class probabilities on $X$}
\label{Sec:webassignment}

\miaojing{
Given the support set $G$ and corresponding labels, 
we begin by propagating
the label information from $G$ to the unlabeled set $X$. For each image $\vx$ in $X$, we use \emph{edge boxes}~\cite{edgeboxes} to extract a collection of $R$ object proposals (regions). Ideally, we would like to have one label per region so we can train an object detector. Since the supervision in our case is very limited, it is not realistic to assign an accurate label per region based only on $G$. Instead, it is more reliable to estimate image-level class probabilities on $X$. Inspired by the two-stream CNN architecture of WSDDN~\cite{wsddn}, we introduce a new way to infer image-level probabilities on $X$, by aggregating region-level class probabilities.
}

\head{Similarity.}
{We} extract a feature vector $\phi(\vr)$ for each region $\vr$ of image $\vx$. We do the same for each image $\vg$ in $G$, extracting a feature vector $\phi(\vg)$. This is a global feature vector.
Let $G_j$ be the support images labeled as class $j$, with $\card{G_j} = k$. Let also $\vr_i$ be the $i$-th region of $\vx$.
We define the $R \times C$ \emph{similarity matrix} $S = \{s_{ij}\}$ with elements
\begin{align}
	s_{ij} \defn \frac{1}{k} \sum_{\vg \in G_j}
	c(\phi(\vr_i), \phi(\vg)),
	\label{eq:sim}
\end{align}
where $c$ denotes cosine similarity.

\head{Voting.}
Inspired by~\cite{wsddn},
{we form $R \times C$ classification matrix $\vcls(S)$ with each row being the softmax of the same row of $S$, implying competition over classes per region; similarly, we form $R \times C$ detection matrix $\vdet(S)$ with each column being the softmax of the same column of $S$, implying competition over regions per class:}
\begin{align}
	\vcls(S)_{ij} \defn \frac{e^{s_{ij}}}{\sum_{j=1}^C e^{s_{ij}}}, \quad
	\vdet(S)_{ij} \defn \frac{e^{s_{ij}}}{\sum_{i=1}^R e^{s_{ij}}}.
	\label{eq:sig}
\end{align}
The $i$-th row of $\vcls(S)$ expresses a vector of class probabilities for region $\vr_i$, while the $j$-th column of $\vdet(S)$ a vector of region probabilities (spatial distribution) for class $j$.

The final image-level class scores $\vsigma(S)$ are obtained by element-wise product of $\vcls(S)$ and $\vdet(S)$ followed by sum pooling over regions
\begin{equation}
	\vsigma(S)_j \defn \sum_{i=1}^R \vcls(S)_{ij} \vdet(S)_{ij}.
	\label{eq:sum}
\end{equation}
Each score $\vsigma(S)_j$ is in $[0,1]$ and can be interpreted as the probability of object class $j$ occurring in image $\vx$.

\head{Discussion.}
The above is a robust \emph{voting strategy} which propagates proposal-level information to the image level, while suppressing noise. Formula~\eq{sim} suggests that region $\vr_i$ will respond for class $j$ if it is similar to any of the support images in $G_j$. While this response is noisy since it is only based on a few examples, it is only maintained if it is among the strongest over all classes and all regions {in an image}.
\miaojing{Note that in~\cite{wsddn}, softmax is applied to two separate data streams 
during learning,
whereas it is applied
to the \emph{same matrix} in our 
work.
}

\miaojing{Alternative ways to transfer label information from $G$ to $X$ would be to directly learn a parametric classifier on $G$ or define a nearest-neighbor classifier on $G$ and infer image-level labels on $X$.
We consider such baselines in our experiments. Their performance is not satisfactory, which highlights the importance of robustly propagating labels from region to image level.
}

\subsection{Teacher and student training}
\label{Sec:teacherstudent}

{Having class probability vectors~(\ref{eq:sum}) per image in $X$,}
a next step would be to convert them to multi-class pseudo-labels and
train the student
directly {on a weakly-supervised detection task}.
Nevertheless, probabilities generated this way rely on the few support images in $G$ for classification, while the object information in the unlabeled set $X$ is not exploited. \miaojing{To further enhance performance, we 
use
\emph{distillation}~\cite{HiVD15,RDG+18} to transfer knowledge between \emph{data} (labeled to unlabeled) and \emph{models} (classification to detection).}
\iavr{In particular, we
distill knowledge from the support set $G$ to the unlabeled set $X$ using a teacher classifier $T$, and then distill this knowledge from the teacher to a student detector $U$.}

\head{Data distillation.}
{We form the teacher $\tea$ as the feature extractor network $\phi$ followed by
a randomly initialized $C$-output fully-connected layer and softmax. We then fine-tune
$\tea$}
on a $C$-way classification task on
$X$. The probabilities~(\ref{eq:sum}) are meant for multi-label classification ($C$ independent binary classifiers), while here we are learning a single $C$-way classifier, {\ie for mutually exclusive labels.}
Given the class probability vector $\vsigma(S)$ for each image $\vx$ in $X$, we take the most likely class
$\arg\max_j \vsigma(S)_j$
as a \emph{$C$-way pseudo-label}. {We fine-tune $\tea$} on these pseudo-labels with a standard cross-entropy loss.

We have also tried several multi-label variants~\cite{wei2016pami,zhu2017cvpr}, which are inferior
to the simple $C$-way cross-entropy loss {in our experiments}. This may be attributed to the class sample distribution in $X$ being unbalanced.

\head{Knowledge distillation.}
{The fine-tuned teacher $\tea$}
encodes object information of $X$ into its network parameters. Directly using its image-level predictions on $X$ would not be appropriate to
{train the student $\stu$}
for detection, because the latter would need multi-class labels. On the other hand, using it as feature extractor to repeat the process of Sec.~\ref{Sec:webassignment} would not make much difference either, as it still produces class probabilities based on $G$. Instead, we use {$\tea$} to \emph{directly classify object proposals in $X$}. Each proposal ideally contains one object, so it is particularly suitable to use {$\tea$} as it was designed: a $C$-way classifier.

Given an input image $\vx$ in $X$, we collect output {class} probabilities of {$\tea$} on each region $\vr_i$ of $\vx$ into a $R \times C$ matrix $A$ with
{element $a_{ij}$ being the probability of class $j$.}
From this matrix, it is possible to estimate new image-level class probabilities by $\vsigma(A)$, similar to~\eq{sum}.
Because it is based on {$\tea$} being trained on $X$ as classifier, while $\vsigma(S)$~\eq{sum} is based on $G$ alone, we combine their strength by averaging both into a probability vector
\begin{equation}
	\hat{\vq} \defn \frac{1}{2} \left(
		\vsigma(S) + \vsigma(A)
	\right)
	\label{eq:avg}
\end{equation}
corresponding to image $\vx$.

An image-level \emph{multi-class pseudo-label} $\hat{\vy} \in \{0,1\}^C$ is then obtained from $\hat{\vq}$ by element-wise thresholding. An element $\hat{y}_j = 1$ specifies that an object of class $j$ occurs in image $\vx$. In the absence of prior knowledge or validation data, we choose $\frac{1}{2}$ as threshold. \iavr{Importantly, an all-negative pseudo-label $\hat{\vq} = (0,\dots,0)$ is possible, \eg when an image does not depict any known class. This simple mechanism allows our method to work in the harder setting where images in $X$ may depict only unknown classes.}

{Those image-level pseudo-labels are all that is needed to obtain an object detector if we use any WSOD pipeline. In particular, we train the student model $\stu$ on weakly-supervised detection on $X$ using \emph{proposal cluster learning} (PCL)~\cite{tang2018pami}. Weakly-labeled images in $G$ are also included into the training with loss weight 1.} 

\head{Inference.}
\miaojing{
At inference, the teacher classifier is not needed. The trained student detector is used directly.
}

%% file: experiment_short.tex
\section{Experiments}
\label{sec:exp}

\subsection{Experimental setup}
\label{Sec:setup}

\head{Unlabeled set $X$.}
We choose the standard object detection datasets PASCAL VOC 2007 and 2012~\cite{pascalvoc} {for the unlabeled set}, having 20 classes. Each dataset contains a \emph{trainval} set and a \emph{test} set. For VOC 2007, the \emph{trainval} set has 5011 images and the \emph{test} set 4952 images. For VOC 2012, the size of \emph{trainval} and \emph{test} sets are 11540 and 10991, respectively.
{We use the \emph{trainval} sets as $X$ to train the object detector by default.  We evaluate the detector on the \emph{test} set.}
{Importantly, except for the support set, we do not use any labels, not even image-level labels in the training set.}

\begin{figure*}[t]
	\centering
	\includegraphics[width=1\textwidth]{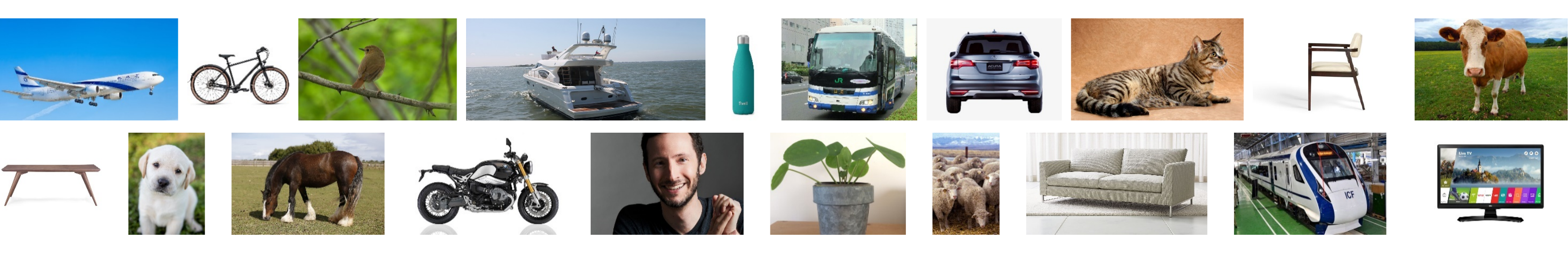} \\[3pt]
	\includegraphics[width=1\textwidth]{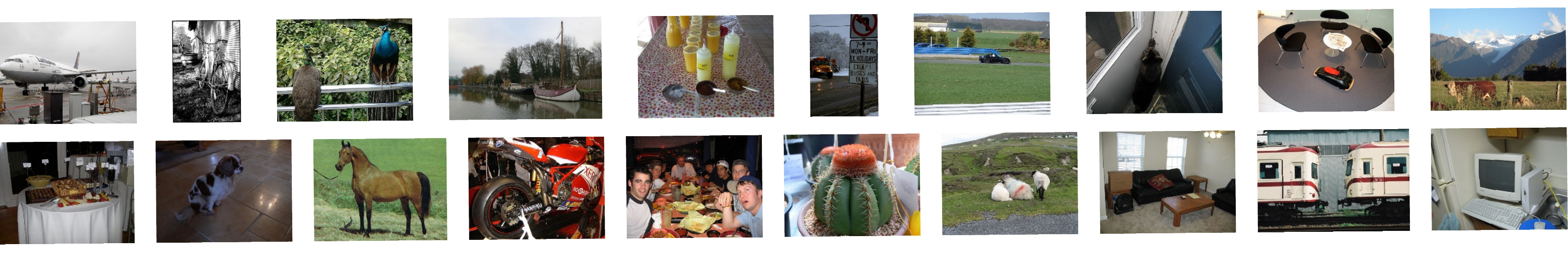}
	\caption{\small (Top) examples of top-ranking web images, using class names as queries. (Bottom) random selection of images from PASCAL VOC 2007.  
	}
	\label{Fig:google}
\end{figure*}

\head{Support set $G$.}
\iavr{Each image in the support set $G$ should depict one of the known $C$ classes (\emph{i.e.} 20 VOC classes). \iavr{A preferable way to collect $G$ is from the web~\cite{modolo2017pami}: we use the class names as text queries and collect the top-$k$ results per class from web image search (\eg~Google).
		The motivation is that these images}
	are clean, \emph{i.e.} they mostly contain objects against a simple background and in a canonical pose and viewpoint, without clutter or occlusion (see examples in Fig.~\ref{Fig:google} (top)). Notwithstanding, they are not perfect, lacking diverse appearance and poses of the object class.
	Collecting images from the web is easy
	and does not need any human effort.
	We choose this option by default.
}

{Another common
	option
	is to randomly
	sample $k$ images per class
	from an existing collection~\cite{ShHX15,marvaniya2012drawing} (\emph{e.g.} VOC 2007). This is a harder setting,
	as these images may depict small objects, multiple instances, object classes in non-canonical pose, clutter and occlusion, \eg bottle, chair, and person in Fig.~\ref{Fig:google} (bottom). }
We experiment with both options.

\head{Networks.} We choose VGG16~\cite{vgg} as our student $\stu$ by default, which is consistent with most WSOD methods~\cite{wsddn,tang2017cvpr,tang2018eccv,shen2019cvpr,tang2018pami}.
{Since the teacher network $T$ (including the feature extractor $\phi$)} is not used at inference time, we choose the more powerful ResNet-152~\cite{resnet}.
Both networks are pre-trained on the ILSVRC classification task~\cite{RDS+14}.

\head{Implementation details.} We use $k=20$ images per class by default for $G$. Following representative WSOD methods~\cite{wsddn,tang2017cvpr,tang2018pami,zhang2018cvpr}, we adopt \emph{edge boxes}~\cite{edgeboxes} to extract $2000$ proposals on average per image in $X$. For the default teacher model $\tea$, we first resize the input image to $256$ pixels on the short side and then crop it to $224 \times 224$. We set the batch size to $128$ and the learning rate to $10^{-3}$ initially with cosine decay. For the default student model $\stu$, we feed the network with one image per batch. The training lasts for $50,000$ iterations in total; the learning rate starts at $10^{-5}$ and decays by an order of magnitude at $35,000$ iterations.

\begin{table*}[t]
	\footnotesize
	\centering
	\setlength{\tabcolsep}{2pt}
	{
		\begin{tabular}{cccccccccccccccccccccc}
			\toprule
			\textsc{Method} & aero & bike & bird & boat & bott & bus & car & cat & char & cow & tabl & dog & hors & mbik & prsn & plat & shep & sofa & tran & tv & mAP \\
			\midrule
			\ours  & 57.9 & {59.7} & 43.2 & {10.5} & 13.1 & 62.7 & 58.6 & 43.9 & 10.6 & 51.1 & \textbf{25.7} & 49.8 & 39.3& {60.6} & 14.9 & 10.9 & {33.5} & 45.2 & {42.5} & \textbf{27.8} &{38.0} \\  
			\ours (07+12) & 51.5& \textbf{65.2} & \textbf {48.9} &  \textbf{13.2} & \textbf{19.7} & \textbf{64.8} & \textbf{59.3} & \textbf{55.5} & \textbf{12.4} & \textbf{59.3} & 24.3 & \textbf{54.1} & \textbf{47.4} & \textbf{62.8} & \textbf{20.7} & \textbf{15.0} & \textbf{39.5} & \textbf{51.3} & \textbf{53.8} & 21.4 & \textbf{42.0} \\
			\midrule
			\iavr{NS-FT} & 56.7 & 37.2 & 31.8 & 10.7 & 4.6 & 44.7 & 42.7 & 51.4 & 3.5 & 17.7 & 4.2 & 37.6 & 22.5 & 51.6 & 13.1 & 10.0 & 28.9 & 36.3 & 39.2 & 14.3 & \iavr{27.9} \\
			\iavr{NS-NN} & 59.2 & 33.3 & 28.3 & 22.5 & 5.4 & 43.7 & 39.3 & 32.3 & 2.3 & 40.1 & 7.5 & 42.2 & 34.2 & 33.2 & 12.6 & 7.7 & 30.5 & 31.1 & 47.6 & 13.7 & \iavr{28.3} \\
			NS-MT-v1& 49.6 & 33.9 & 29.6 & 15.5 & 9.5 & 47.9 & 32.9 & 49.1 & 0.2 & 13.2 & 21.1 & 34.4 & 19.7 & 31.5 & 9.6 & 9.9 & 35.6 & 43.1 & 38.9 & 15.0 & 27.0\\
			NS-MT-v2 & 46.6 & 22.5 & 25.6 & 7.4 & 4.2 & 49.0 & 35.4 & 71.4 & 0.4 & 25.0 & 22.5 & 56.7 & 38.3 & 58.8 & 6.9 & 10.3 & 27.0 & 59.1 & 22.9 & 6.0 & 29.8 \\
			\midrule
			WSDDN~\cite{wsddn}& 39.4 &  50.1 & 31.5 & 16.3 & 12.6& 64.5  & 42.8 & 42.6 & 10.1 & 35.7 &  24.9 &  38.2 &  34.4 &  55.6 &  9.4  & 14.7 &  30.2 & 40.7& 54.7 & 46.9 & 34.8 \\
			OICR~\cite{tang2017cvpr} & \textbf{58.0} & 62.4 & 31.1 & 19.4  & 13.0  & \textbf{65.1} & 62.2 & 28.4 &  24.8& 44.7 & 30.6& 25.3 & 37.8 & 65.5 & 15.7 &24.1 &41.7& 46.9& \textbf{64.3}& 62.6& 41.2\\
			WSRPN~\cite{tang2018eccv} & 57.9 & \textbf{70.5}& 37.8 &5.7 &21.0& 66.1& \textbf{69.2}& 59.4 &3.4 &\textbf{57.1}& \textbf{57.3}& 35.2& \textbf{64.2}& \textbf{68.6}& \textbf{32.8}& \textbf{28.6}& \textbf{50.8}& 49.5& 41.1& 30.0& 45.3 \\
			PCL~\cite{tang2018pami} & 54.4& {69.0} & 39.3& 19.2& 15.7& 62.9& 64.4& 30.0& \textbf{25.1}& 52.5& 44.4& 19.6 & 39.3& 67.7 & 17.8& 22.9& 46.6& \textbf{57.5} & 58.6& \textbf{63.0} & 43.5\\
			WS-JDS~\cite{shen2019cvpr} & 52.0 & 64.5 & \textbf{45.5} & \textbf{26.7} & \textbf{27.9}& 60.5& 47.8& \textbf{59.7}& 13.0 & 50.4& 46.4& \textbf{56.3}& 49.6& 60.7 & 25.4 & 28.2 & 50.0 & 51.4& 66.5& 29.7 & \textbf{45.6}\\
			\bottomrule
		\end{tabular}
	}
	\vspace{3pt}
	\caption{Detection mAP on the \emph{test} set of PASCAL VOC 2007. \ours: our nano-supervised object detection framework;
		\iavr{NS-FT: nano-supervised fine-tuning; NS-NN: nano-supervised nearest neighbor; NS-MT: Nano-supervised mean teacher}.
		Unless otherwise stated, \ours, NS-FT, NS-NN use $k = 20$ support images per class by default. All compared methods~\cite{wsddn,tang2017cvpr,tang2018eccv,tang2018pami,shen2019cvpr} use the image-level labels in the unlabeled set $X$; \ours,
		\iavr{NS-FT, NS-NN and NS-MT}
		do not.}
	\label{tab:det_map_voc2007}
\end{table*}

\yannis{
	\head{Evaluation protocol.}
	We evaluate the performance of our NSOD framework
	on both image classification and object detection.
	For image classification, we measure the \emph{average precision} (AP) and \emph{mean AP} (mAP) for \emph{multi-class predictions}~\cite{wei2016pami,zhu2017cvpr}, as well as the accuracy of the top-1 class prediction per image on the \emph{trainval} set of $X$.
	For object detection, we quantify
	localization performance
	on the \emph{trainval} set by \emph{CorLoc}~\cite{wsddn,shi2016eccv,tang2017cvpr,zhang2018cvpr} and detection performance on the \emph{test} set by mAP.
	At test time, the detector can localize multiple instances of the same class per image and mAP is identical to what is used to evaluate fully supervised object detectors with an IoU threshold of 0.5. By using the same IoU threshold, we measure the recall rate of \emph{edge boxes} over ground truth to be 92.51\% and 91.27\% on VOC 2007 and 2012 respectively. This shows the capacity of \emph{edge boxes} to cover most ground truth regions.
	
	\head{Evaluation scenarios.}
	Below, we first present the object detection results on the \emph{test} set of $X$ under two scenarios: \emph{support set $G$ by web search} (\autoref{sec:web}) and \emph{by sampling VOC 2007} (\autoref{sec:voc}). Then in the ablation study (\autoref{sec:ablation}), we provide detection and classification results on the \emph{trainval} set of $X$ using the web search scenario.
}

\subsection{{Support set $G$ by web search}}
\label{sec:web}
{We first collect the support set by web search and evaluate our NSOD on both VOC 2007 and 2012. We also combine the two sets as well as images from ImageNet as distractors to evaluate our method in the wild.}

\subsubsection{Results on VOC 2007}\label{sec:results-voc2007}
\head{Comparison to weakly-supervised methods.}
We compare to several representative WSOD methods~\cite{wsddn,tang2017cvpr,tang2018eccv,tang2018pami,shen2019cvpr} in Table~\ref{tab:det_map_voc2007}. For fair comparison, all these methods use the same VGG16 backbone as we do, without bells and whistles.
\ours requires no annotation on the unlabeled set $X$, while weakly-supervised methods assume image-level labels for all images in $X$.

\begin{figure*}[t]
	\centering
	\includegraphics[width = 1.0\textwidth]{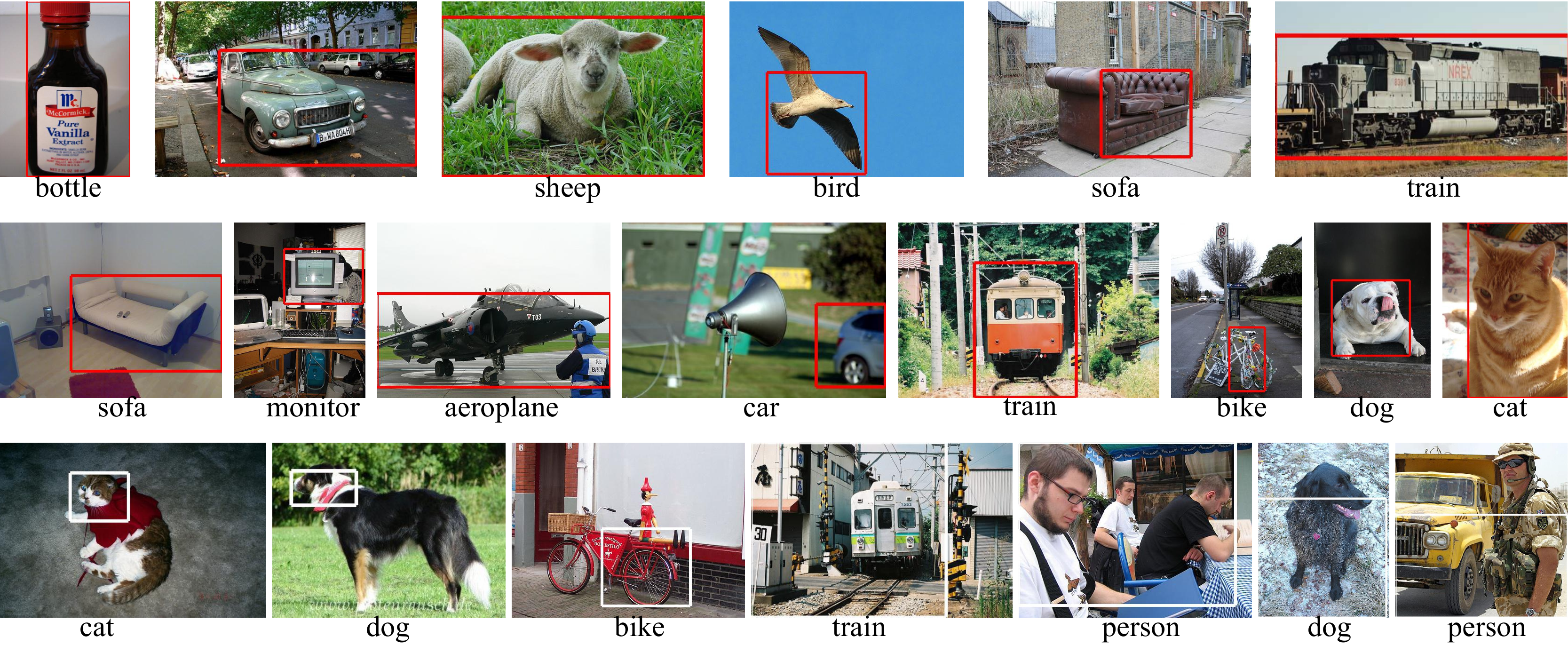}
	\caption{\small Detection results of \ours on PASCAL VOC 2007, using default settings ($k=20$).
		Top 2 rows: positive results (red boxes). Bottom row: failure cases (white boxes).
	}
	\label{Fig:result}
\end{figure*}

One directly competing method is PCL trained on gro\-und truth image-level labels in $X$. Despite using no annotation on $X$, \ours achieves an mAP that is only 5.5\% below that of PCL (38.0 \vs 43.5). The result is is also competitive to other methods, \eg OICR~\cite{tang2017cvpr}, WSRPN~\cite{tang2018pami}. {There are also WSOD methods employing large-scale web images/videos as extra data. For instance, \cite{tao2018tmm} and \cite{singh2019cvpr} build on the WSDDN pipeline~\cite{wsddn} and produce mAP 36.8 and 39.4 on VOC 2007, respectively. Unlike these works, our \ours uses few web images, an unlabeled set, and an advanced WSOD pipeline. Importantly, \ours also delivers competitive mAP.}
Fig.~\ref{Fig:result} gives some examples of detection results of \ours on PASCAL VOC 2007.

\head{Comparison to semi-supervised methods.}
\iavr{
	We first compare NSOD to two semi-supervised baselines: (1) fine-tune the teacher $T$ on $G$ as a $C$-way classifier and use it to make predictions on $X$, referred to as NS-FT; (2) use $\phi$ from $T$ as a global feature extractor to find the nearest neighbor in $G$ for each image in $X$, referred to as NS-NN. In both cases, we use the same support set with NSOD, infer image-level $C$-way pseudo-labels on $X$ and use them to train the student $U$ by PCL. As shown in Table~\ref{tab:det_map_voc2007}, NS-FT and NS-NN deliver a mAP of 27.9 and 28.3, respectively.} \miaojing{Comparing to the mAP 38.0 of NSOD on the same setting ($k=20$), these baselines are not satisfactory.
	This is due to the limited the supervision from the support set and justifies our choice of propagating labels from region to image level. }

We then adapt the \emph{mean teacher}~\cite{TV17} semi-supervised classification method to our setting. We use it in two ways: 1) using image-level class probabilities $\hat \vq$~ in Eq.(\ref{eq:avg}), we select the top-$z$ scored images as positive for each class and the rest we treat as negative.
With those pseudo-labels, we train PCL on VGG16, applying the consistency loss of~\cite{TV17} to image-level predictions on $X$. We call this \emph{nano-supervised mean teacher - variant 1} (NS-MT-v1).
We choose $z = 300$ as it works the best in practice. NS-MT then yields an mAP of 27.0 as shown in Table~\ref{tab:det_map_voc2007}. This result is lower than our \ours by {11.0\%} ({27.0} \vs 38.0); 2) using the labeled support set $G$ and unlabeled set $X$, we train the teacher model $T$ as a mean-teacher by applying the cross-entropy loss on G and consistency loss between G and X. The rest pipeline remains as in NOSD. We name this \emph{nano-supervised mean teacher - variant 2} (NS-MT-v2) and obtain an mAP of 29.8 \vs 38.0 of NSOD. Both variants of NS-MT are clearly inferior to NSOD, which suggests that it is not straightforward to transfer a successful semi-supervised approach from the classification to the detection task.

We have also tried to directly infer object bounding
boxes on the \emph{test} set of VOC 2007 using naive approaches.
In particular:
\begin{enumerate}
	\item We use the $C$-way classifier $T$ trained on $X$ (stage 3) to directly predict the class probabilities of object proposals per image in the test set.
	\item We adopt a naive $k$-NN classifier by computing the feature similarities from images in the support set $G$ to the object proposals per image in the test set.
\end{enumerate}
The mAP in both cases can be measured by ranking the proposals by their class probabilities. These methods fail, producing mAP lower than 10. We should emphasize the importance
of propagating similarity scores from region-level to image-level as we do in NSOD.

\subsubsection{Results on VOC 2012}
Using the same support set $G$, we train an object detector with our \ours on VOC 2012. The mAP is reported on the \emph{test} set of VOC 2012 and compared to representative WSOD methods~\cite{tang2017cvpr,tang2018pami,zhang2018cvpr} in Table~\ref{tab:det_corloc_voc2012}.
Despite not using any VOC 2012 labels, \ours is only 4.0\% below PCL (36.6 \vs 40.6).

\begin{table*}[t]
	\footnotesize
	\centering
	\setlength{\tabcolsep}{2pt}
	{
		\begin{tabular}{cccccccccccccccccccccc}
			\toprule
			\textsc{Method} & aero & bike & bird & boat & bott & bus & car & cat & char & cow & tabl & dog & hors & mbik & prsn & plat & shep & sofa & tran & tv & mAP \\
			\midrule
			OICR~\cite{tang2017cvpr} & \textbf{67.7} & 61.2 &  41.5 &  \textbf{25.6} &  22.2 &  54.6 &  49.7 &  25.4 &  \textbf{19.9} &  47.0 &  18.1 &  26.0 &  38.9 &  67.7 &  2.0 & 22.6 &  41.1 &  34.3 &  37.9 & 55.3 &  37.9\\
			ZLDN~\cite{zhang2018cvpr}& 54.3 & 63.7 & 43.1 & 16.9 & 21.5 &  \textbf{57.8}&  \textbf{60.4} &  50.9&  1.2 & 51.5&  \textbf{44.4} &  36.6 & \textbf{ 63.6} &  59.3& 12.8& 25.6& \textbf{47.8}& \textbf{47.2}& 48.9& 50.6& \textbf{42.9} \\
			PCL~\cite{tang2018pami} & 58.2 & \textbf{66.0} & 41.8 & 24.8 & \textbf{27.2}& 55.7 & 55.2& 28.5 & 16.6& 51.0 & 17.5 & 28.6& 49.7& \textbf{70.5}& 7.1& \textbf{25.7}& 47.5 & 36.6 & 44.1& \textbf{59.2} & 40.6 \\
			\midrule
			\ours & 56.3 & 27.6 & 42.2 & 10.9 &  23.8 & 55.1 & 46.2 & 36.6  & 5.6  & 51.8 & 15.5 & 55.9 & 54.0 & 63.6 & \textbf{23.5} & 10.8 & 43.1 & 39.2 & \textbf{49.0} & 21.5 & 36.6 \\
			\ours (07+12) &  57.3 & 50.7 & \textbf{49.2} & 11.3 & 21.2 & 56.8 & 46.4 & \textbf{55.0} &  6.6 & \textbf{52.7} & 12.8 & \textbf{61.8} & 45.8 & 64.7 & 18.9 & 10.5 & 34.9 & 41.0 & 48.1 & 19.9 & 38.6\\
			\bottomrule
		\end{tabular}
	}
	\vspace{3pt}
	\caption{Detection mAP on \emph{test} set of PASCAL VOC 2012. Our \ours uses $k = 20$ support images per class. All compared methods~\cite{tang2017cvpr,tang2018pami,zhang2018cvpr} use the image-level labels in the unlabeled set $X$; our \ours does not.}
	\label{tab:det_corloc_voc2012}
\end{table*}

\begin{table*}[t]
	\footnotesize
	\centering
	\setlength{\tabcolsep}{2pt}
	{
		\begin{tabular}{cccccccccccccccccccccc}
			\toprule
			\textsc{Method} & aero & bike & bird & boat & bott & bus & car & cat & char & cow & tabl & dog & hors & mbik & prsn & plat & shep & sofa & tran & tv & mAP \\
			\midrule
			{\ours (07+Dis5k)}   & 59.3 & 35.4 & 37.6 & 16.6 & 7.5 & 59.1 & 59.0 & 42.2 & 9.0 & 47.4 & 33.2 & 50.8 & 46.3 & 52.4 & 15.1 & 18.7 & 44.2 & 50.3 & 51.6 & 35.3 & 37.6\\
			{\ours (07+Dis10k)}  & 56.5 & 36.0 & 34.6 & 12.7 & 5.7 & 56.6 & 56.2 & 40.1 & 8.5 & 44.9 & 31.1 & 46.0 & 41.6 & 55.1 & 15.7 & 15.1 & 39.9 & 46.8 & 47.6 & 31.2 & 36.5 \\
			{\ours (07+Dis15k)} & {57.1} & {56.4} & {18.5} & {17.2} & {15.1} & {62.0} & {58.3} & {44.6} & {8.4} & {41.8} & {30.0} & {49.1} & {40.8} & {59.7} & {19.1} & {12.3} & {29.3} & {34.3} & {36.7} & {29.1} & {36.0} \\
			{\ours (07+Dis20k)} & {59.0} & {46.7} & {21.0} & {16.7} & {11.0} & {61.0} & {57.9} & {56.1} & {10.6} & {38.0} & {32.0} & {53.1} & {44.0} & {57.3} & {16.3} & {13.4} & {29.4} & {23.4} & {35.4} & {30.6} & {35.7} \\
			\ours (07+12+Dis5k)  & 59.8 & 65.8 & 50.1 & 12.5 & 16.5 & 58.6 & 52.1 & 57.0 & 15.8 & 51.1 & 31.5 & 53.9 & 36.4 & 58.8 & 18.1 & 15.4 & 43.3 & 50.4 & 48.1 & 38.8 & {41.7} \\
			\ours (07+12+Dis10k) & 51.4 & 68.1 & 36.1 & 11.8 & 17.7 & 59.6 & 63.1 & 61.8 & 10.2 & 46.5 & 32.1 & 57.0 & 37.1 & 61.3 & 17.7 & 17.1 & 44.0 & 47.7 & 44.9 & 33.0 & {40.9} \\
			{\ours (07+12+Dis15k)} & {54.7} & {52.2} & {29.0} & {18.7} & {18.4} & {63.6} & {60.2} & {44.4} & {8.9} & {57.1} & {29.0} & {58.7} & {49.2} & {60.8} & {20.3} & {13.0} & {44.1} & {48.7} & {44.9} & {38.1} & {40.7} \\
			{\ours (07+12+Dis20k)} & {59.2} & {53.6} & {33.7} & {12.3} & {18.6} & {59.6} & {55.3} & {44.3} & {9.7} & {50.8} & {35.3} & {50.4} & {53.5} & {58.7} & {22.3} & {15.4} & {39.4} & {45.8} & {40.4} & {42.9} & {40.2} \\
			\bottomrule
		\end{tabular}
	}
	\vspace{3pt}
	\caption{Detection mAP on the \emph{test} set of PASCAL VOC 2007 in the presence of distractors. \ours: our object detection framework.}
	\label{tab:det_map_ablation}
\end{table*}

\subsubsection{Results on VOC 2007 + 2012}\label{sec:Results-2007-2012}
Because $X$ is unlabeled and our method is computationally efficient, we can easily improve performance by simply using more unlabeled data.
As shown in Table~\ref{tab:det_map_voc2007} and~\ref{tab:det_corloc_voc2012}, if we train \ours on the union of VOC 2007 and VOC 2012 (07+12) on a large-scale, the mAP can be further improved on the \emph{test} set of both VOC 2007 and 2012. For instance, on VOC 2007, NSOD (07+12) yields a mAP of 42.0, which is an improvement by +4\% over using VOC 2007 alone. Since neither set is labeled, this improvement comes at almost no cost. This result is only 1.5\% below PCL (42.0 \vs 43.5), 
and even outperforms WSDDN~\cite{wsddn} and OICR~\cite{tang2017cvpr} when trained on VOC 2007 with image-level labels. This is a strong result that confirms the value of our core contribution; similarly, on VOC 2012, NSOD (07+12) increases the mAP to 38.6, now outperforming OICR.

\subsubsection{Results on PASCAL VOC + \yannis{Distractors}}
\miaojing{
	Despite being used without labels, VOC 2007 and 2012 are still \emph{curated}, \ie images depict at least one of the target classes. To further validate the effectiveness of our methd, we experiment with unlabeled data in the wild for $X$, \ie, using images depicting unknown rather than target classes. In particular, we randomly select 5k, 10k, 15k and 20k images from ImageNet~\cite{RDS+14} and use the union of this set and VOC 2007, denoted by 07+Dis5k, 07+Dis10k,  07+Dis15k, and  07+Dis20k, as $X$. Although there may be overlap between the 1000 ImageNet classes and the 20 PASCAL VOC classes, these images mostly contain unknown classes and play the role of distractors. The evaluation is on the test set of VOC 2007. As shown in Table~\ref{tab:det_map_ablation}, 07+Dis5k yields a mAP of 37.6, which almost retains the performance of using VOC 2007 alone as $X$ (38.0). Further increasing the distractor set causes very little performance drop. For instance, the mAP for NSOD (07+Dis15k) on VOC 2007 is 36.0, which is -0.5\% compared to that of NSOD (07+Dis10k); while for NSOD (07+Dis20k), only -0.3\% is further observed upon NSOD (07+Dis15k). Considering the unlabelled set of VOC 2007 plus distractors as a whole, this shows our method is able to discover the relevant data and filter out most of the distractors, despite the distractor set being much larger than the curated set.}

Furthermore, we add the union of 5k/10k/15k/20k images from ImageNet and 10k images from VOC 2012 to VOC 2007, denoted by 07+12+Dis5k/10k/15k/20k, which achieves mAP 41.7/40.9/40.7/40.2. Similar to above, the performance drop by adding more distractors is also very small. These results are only slightly lower than that of NSOD (07+12) (mAP 42.0, Table 1), indicating our method mostly ignores distractors. We find that the distractors are mostly assigned no pseudo-labels due to thresholding of $\hat{\vq}$~\eq{avg} in \ours. In other words, within the additional noisy unlabeled data (12+Dist5k,10k,15k,20k), our method discovers the relevant data (12) and uses them to improve from mAP 38.0 (07 alone), while mostly ignoring the irrelevant data (\ie Dis5k,10k,15k,20k). The additional noisy unlabeled data is meant to represent data in the wild, which can be obtained for free. Hence, depending on the ratio of labelled to unlabelled data, it is possible to improve the detection performance with no annotation cost.

\begin{figure}[t]
	\centering
	\includegraphics[width = 0.9\linewidth]{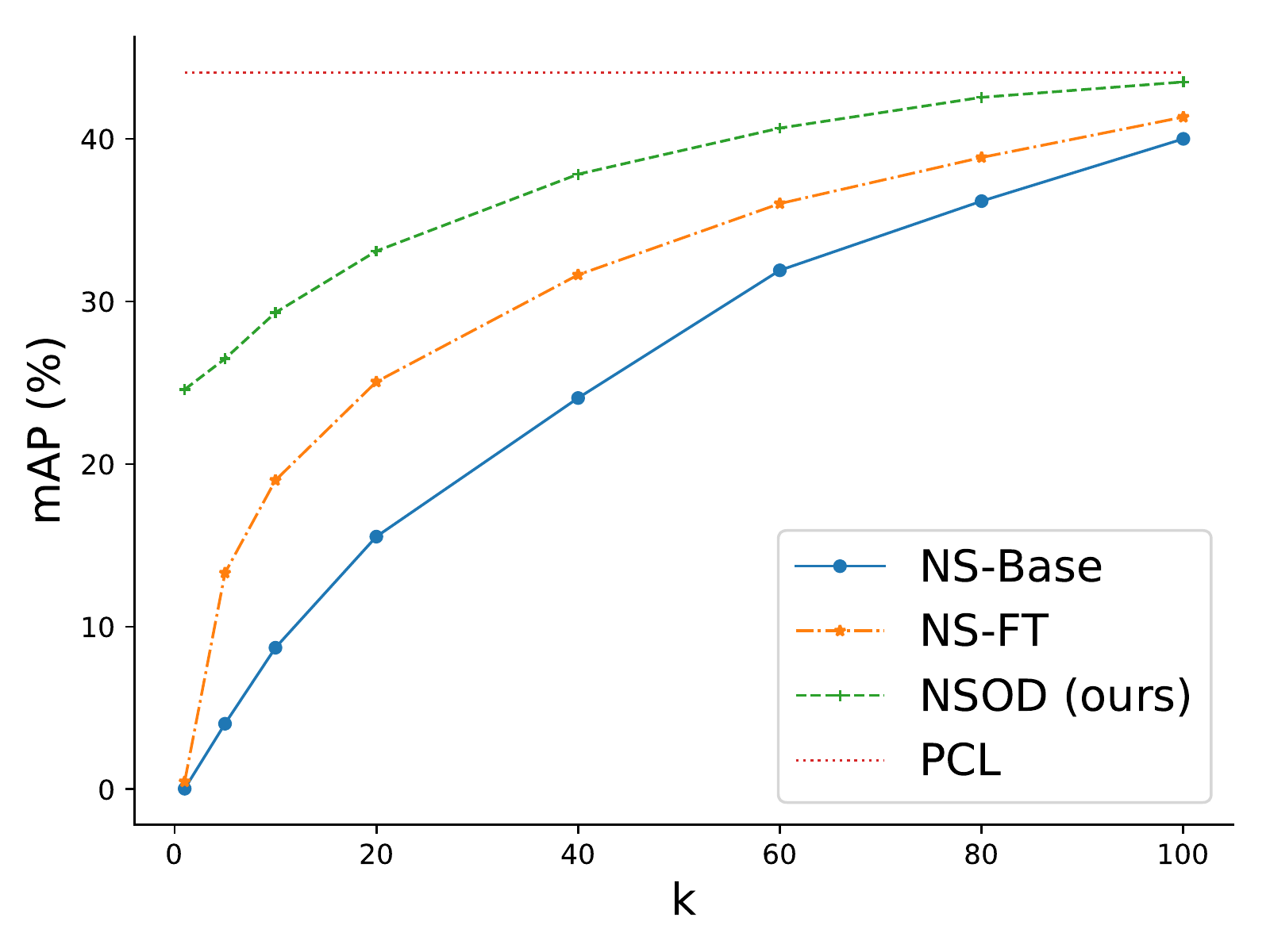}
	\caption{\small Detection mAP of \ours, NS-Base, NS-FT and PCL on PASCAL VOC 2007, using different number $k$ of images per class as support set.
	}
	\label{Fig:result_voc_support}
\end{figure}

\begin{table*}
	\footnotesize
	\centering
	\setlength{\tabcolsep}{2pt}
	{
		\begin{tabular}{cccccccccccccccccccccc}
			\toprule
			\textsc{Method} & aero & bike & bird & boat & bott & bus & car & cat & char & cow & tabl & dog & hors & mbik & prsn & plat & shep & sofa & tran & tv & mAP \\
			\midrule
			\oursg & 88.8 & 85.8 & 98.0 & 67.8 & 79.4 & 68.4 & \textbf{96.8} & 95.1 & \textbf{80.6} & 72.1 & 38.9 & 93.4 & \textbf{82.3} & 65.2 & 98.0 & 56.7 & 70.1 & 55.6 & 72.0 & 60.2 & 76.3\\
			\oursx  & 86.4 & \textbf{96.9} & 97.1 & \textbf{71.4} & \textbf{98.5} & 67.1 & 89.9 & 95.1 & 80.0 & 66.8 & 36.5 & 92.9 & 74.2 & 62.9 & 96.9 & 53.1 & 59.9 & 58.8 & 70.1 & \textbf{78.9} & 76.7\\
			\ours & \textbf{91.2} & 90.7 & \textbf{98.0} & 71.1 & 94.3 & \textbf{73.8} & 95.8 & 95.5 & 80.5 & \textbf{74.7} & \textbf{39.1} & \textbf{95.3} & 81.2 & \textbf{66.9} & \textbf{98.4} & \textbf{58.7} & \textbf{73.8} & \textbf{59.7} & \textbf{75.6} & 70.4 & \textbf{79.2}\\
			\midrule
			\textsc{Method} & aero & bike & bird & boat & bottle & bus & car & cat & chair & cow & table & dog & horse & mbike & persn & plant & sheep & sofa & train & tv & mAcc \\
			\midrule
			\oursg & 92.2 & \textbf{97.7} & 99.1 & 78.7 & 100.0 & 73.0 & 93.2 & \textbf{98.5} & \textbf{89.2} & 82.0 & 41.8 & \textbf{97.7} & 77.7 & 72.0 & \textbf{99.7} & 63.7 & 68.7 & 63.0 & 77.5 & 87.5 & 82.7\\
			\oursx &  93.1 & 93.4 & 98.2 & 79.2 & 100.0 & 78.9 & 96.3 & 96.7 & 84.0 & 83.5 & 45.1 & 95.7 & 84.2 & 72.9 & 98.5 & 73.3 & 77.2 & 66.1 & 83.3 & \textbf{87.7} & 84.3\\
			\ours & \textbf{93.8} & 92.4 & \textbf{99.3} & \textbf{80.4} & \textbf{100.0} & \textbf{81.1} & \textbf{97.8} &{97.1} & 78.6 & \textbf{86.7} & \textbf{49.7} & 97.1 & \textbf{88.2} & \textbf{77.2} & 99.6 & \textbf{79.7} & \textbf{79.1} & \textbf{67.9} & \textbf{87.5} & 85.6 & \textbf{85.9} \\
			\bottomrule
		\end{tabular}
	}
	\vspace{3pt}
	\caption{Classification mAP for multi-class prediction (top) and classification mAcc for top-1 class prediction (bottom) on the \emph{trainval} set of PASCAL VOC 2007. \ours: our Nano-supervised object detection framework.}
	\label{tab:cls_map_voc2007_ablation}
\end{table*}

\begin{table*}[t]
	\footnotesize
	\centering
	\setlength{\tabcolsep}{2pt}
	{
		\begin{tabular}{cccccccccccccccccccccc}
			\toprule
			\textsc{Method} & aero & bike & bird & boat & bott & bus & car & cat & char & cow & tabl & dog & hors & mbik & prsn & plat & shep & sofa & tran & tv & mAP \\
			\midrule
			\oursg & 57.2 & 52.7 & 36.0 & 14.1 & 11.0 & 50.6 & 46.9 & 35.8 & 5.7 & 47.1 & 16.1 & 52.8 & 34.3 & 54.4 & 14.8 & 11.4 & 29.0 & 48.8 & 43.4 & 13.9 & 33.9 \\
			\oursx & \textbf{58.5} & 51.5 & 37.5 & 11.6 & 10.6 & 55.3 & 48.2 & 40.4 & 5.8 & 49.9 & 16.0 & 51.3 & 31.6 & 56.3 & 14.6 & 9.0 & 34.3 & 45.5 & 42.2 & 20.3 & 34.5\\
			\ours  & 57.9 & {59.7} & 43.2 & {10.5} & 13.1 & 62.7 & 58.6 & 43.9 & 10.6 & 51.1 & \textbf{25.7} & 49.8 & 39.3& {60.6} & 14.9 & 10.9 & {33.5} & 45.2 & {42.5} & \textbf{27.8} &{38.0} \\  
			\midrule
			\ours ($k=1$) & 53.0 & 58.0 & 24.4 & 13.3 & 11.3 & 41.3 & 43.8 & 43.6 & 2.3 & 50.3 & 6.1 & 32.4 & 19.0 & 50.5 & 15.0 & 8.7 & 35.7 & 41.7 & 42.8 & 6.2 & 30.0
			\\
			\ours ($k=10$) & 57.2 & 27.8 & 40.4 & 9.7 & 11.2 & 61.2& 57.0 & 25.9 & 13.4 & 47.2 & 6.2 & 45.5 & 35.7 & 53.0 & 21.2 & 14.1 & 34.8 & 43.7 & 39.8 & 19.8 & 33.2\\
			\ours ($k=20$)  & 57.9 & {59.7} & 43.2 & {10.5} & 13.1 & 62.7 & 58.6 & 43.9 & 10.6 & 51.1 & \textbf{25.7} & 49.8 & 39.3& {60.6} & 14.9 & 10.9 & {33.5} & 45.2 & {42.5} & \textbf{27.8} &{38.0} \\ 
			\bottomrule
		\end{tabular}
	}
	\vspace{3pt}
	\caption{\emph{Ablation study}. Detection mAP on the \emph{test} set of PASCAL VOC 2007. \ours: our nano-supervised object detection framework.}
	\label{tab:det_map_voc2007_ablation}
\end{table*}

\begin{table*}[!]
	\footnotesize
	\centering
	\setlength{\tabcolsep}{2pt}
	{
		\begin{tabular}{cccccccccccccccccccccc}
			\toprule
			\textsc{Method} & aero & bike & bird & boat & bott & bus & car & cat & char & cow & tabl & dog & hors & mbik & prsn & plat & shep & sofa & tran & tv & mAP \\
			\midrule
			WSDDN~\cite{wsddn}& 65.1 & 58.8 & 58.5 & 33.1 & 39.8 & 68.3 & 60.2 & 59.6&  34.8 & 64.5&  30.5&  43.0&  56.8&  82.4& 25.5& 41.6& 61.5& 55.9& 65.9& 63.7& 53.5 \\
			OICR~\cite{tang2017cvpr} & 81.7 & 80.4& 48.7 & \textbf{49.5} & 32.8&  81.7 & 85.4 & 40.1 & \textbf{40.6}&  79.5 &  35.7 &  33.7 &  60.5&  88.8 &  21.8&  57.9&  76.3&  59.9&  75.3&  \textbf{81.4}&  60.6\\
			WSRPN~\cite{tang2018eccv} & 77.5 &{81.2}& 55.3& 19.7& 44.3 &80.2& \textbf{86.6} &\textbf{69.5} &10.1 &\textbf{87.7} &\textbf{68.4} &52.1 &\textbf{84.4}& \textbf{91.6}& \textbf{57.4} &\textbf{63.4} &\textbf{77.3} &58.1 &57.0 &53.8 &63.8 \\
			PCL~\cite{tang2018pami} & 79.6 & \textbf{85.5}& 62.2 & 47.9 & 37.0 & \textbf{83.8 }& {83.4} & 43.0 & 38.3 &  80.1 & 50.6 & 30.9&  57.8 &  90.8&  27.0& 58.2& 75.3& \textbf{68.5}& 75.7& 78.9& 62.7\\
			WS-JDS~\cite{shen2019cvpr} & \textbf{82.9} & 74.0& \textbf{73.4}& 47.1& \textbf{60.9}& 80.4& 77.5& 78.8& 18.6& 70.0 &56.7& 67.0 &64.5& 84.0& 47.0& 50.1& 71.9& 57.6 &\textbf{83.3} &43.5& \textbf{64.5}\\
			\midrule
			\ours & 80.0 & 73.3 & 66.1& 34.0 & 29.0& 72.6& 76.5& 56.4 & 17.7& 74.7& 47.5& 61.4& 60.5& 86.4& 31.9& 36.6& 60.8& 59.1 & 57.4& 49.1& 56.6\\
			\ours (07+12) & 78.3 & 78.4 & 70.3 & 34.0 & 34.0 & 75.1 & 76.6 & 66.9 & 24.8 & 76.0 & 45.6 & \textbf{69.8} & 67.7 & 88.8 & 34.4 & 41.4 & 67.0 & 62.1 & 67.3 & 40.9 & 60.0\\
			\bottomrule
		\end{tabular}
	}
	\vspace{3pt}
	\caption{CorLoc on the \emph{trainval} set of PASCAL VOC 2007. All compared methods~\cite{wsddn,tang2017cvpr,tang2018eccv,tang2018pami,shen2019cvpr} use the image-level labels in $X$; our \ours does not.}
	\label{tab:det_corloc_voc2007}
\end{table*}

\subsection{{Support set by sampling VOC 2007}}
\label{sec:voc}

{As discussed in Sec.~\ref{Sec:setup}, the support set $G$ can be collected by randomly selecting $k$ images per class from the unlabeled set $X$. This is more challenging than web search, as one image may depict more than one object, as shown in Fig.~\ref{Fig:google}. We randomly sample $k \in \{1, 5, 10, 20, 40, 60, 80, 100\}$ images per class from
	VOC 2007 with image-level labels as $G$ and evaluate on its \emph{test} set.} 
We compare \ours with two baselines: (1) only using $G$ to train the student $U$, denoted by NS-Base; (2) using NS-FT as described in Sec.~\ref{sec:results-voc2007}.

{As shown in Figure~\ref{Fig:result_voc_support}, \ours yields significantly higher mAP at every $k$ compared to the baselines. In particularly, with small $k$, our improvement is substantial; with $k=80$ (around 30\% of VOC 2007 training data), \ours achieves accuracy already very close (on par) to PCL~\cite{tang2018pami} (dotted horizontal line) that uses image-level labels of 100\% data in VOC 2007.}

\subsection{Ablation Study}
\label{sec:ablation}
\yannis{
	We conduct the ablation study
	on our labeling strategy, support set size, and localization on the \emph{trainval} set of PASCAL VOC 2007. The support set $G$ is collected by web search.
	
	\subsubsection{Labeling strategy (classification)}
	Referring to \autoref{Sec:teacherstudent} in the paper, we ablate combining $\vsigma(S)$ and $\vsigma(A)$ to generate image-level pseudo-labels. $\vsigma(S)$ is computed based on $G$ alone, while $\vsigma(A)$ is computed based on the teacher model trained on $X$. We apply a hard threshold of $\frac{1}{2}$ on the predicted class probabilities of $\vsigma(S)$ and $\vsigma(A)$ to generate two sets of image-level pseudo-labels. We train two different models separately on the two sets of pseudo-labels, which we denote by \oursg and \oursx, respectively.
	
	The classification accuracy of the two sets of pseudo-labels is first evaluated on the \emph{trainval} set of VOC 2007 and shown in Table~\ref{tab:cls_map_voc2007_ablation}. It can be seen that \oursg and \oursx produce a similar classification mAP of $76.3$ \vs $76.7$, while the AP on individual classes differs. However, in terms of top-1 class accuracy, \oursx is better than \oursg. This is reasonable, as \oursx is fine-tuned as a $C$-way classifier, which takes the top-1 class predictions of $\vsigma(S)$ as pseudo-labels. The two sets of pseudo-labels are complementary by averaging $\vsigma(S)$ and $\vsigma(A)$ according to~Eq.(4), denoted by \ours. This improves both multi-class and top-1 class predictions, reaching the highest scores of $79.2$ and $85.9$, respectively.
	
	\subsubsection{Labeling strategy (detection)}
	To further investigate the complementary effect of the two models \oursg and \oursx, we evaluate their detection result on the \emph{test} set of VOC 2007 (Table~\ref{tab:det_map_voc2007_ablation}). The mAP of \oursx (34.5) is slightly greater than that of \oursg (33.9). Their combination (our full model \ours) further increases mAP by $+3.5\%$ to $38.0$.
	The detection result on the \emph{test} set is consistent with the classification result on the \emph{trainval} set, which validates our idea of distilling knowledge from the support set to the unlabeled set and from the teacher to the student model.
	
	\subsubsection{Support set size} \label{sec:support_set_size}
	We evaluate performance for different number $k$ of web images per class of the support set $G$ in Table~\ref{tab:det_map_voc2007_ablation}: mAP is 30.0 for $k=1$, 33.2 for $k=10$ and 38.0 for $k=20$.
	Further increasing $k$ presumably brings more noisy examples.  How to deal with large-scale noisy web images/videos is an open problem~\cite{guo2018eccv,tao2018tmm,singh2019cvpr,liang2015cvpr}. We keep $G$ small to avoid bringing too many noisy images, while at the same time using the unlabeled unlabeled set $X$ for
	{more diversity.}
}

\yannis{
	\subsubsection{Localization on the trainval set}
	Apart from the mAP on the \emph{test} set, in Table~\ref{tab:det_corloc_voc2007} we report CorLoc on the \emph{trainval} set of VOC 2007, as is common for weakly-supervised detection methods~\cite{wsddn,tang2017cvpr,tang2018eccv,tang2018pami,shen2019cvpr}. Our NSOD delivers CorLoc 56.6, which is very close to other WSOD methods despite using no annotations on $X$. 
	Like in \autoref{sec:Results-2007-2012}, if we train \ours on the union of VOC 2007 and VOC 2012 (07+12) on a large-scale, the CorLoc of \ours(07+12) on VOC 2007 (see Table~\ref{tab:det_corloc_voc2007}) is increased to 60.0, which is only 2.7\% below PCL (62.7) and generally among the best-performing WSOD methods (\eg OICR has 60.6).
}

%% file: discussion.tex
\section{Discussion}
\label{sec:discussion}

Our nano-supervised object detection framework basically begins with a combination of \emph{few-shot} and \emph{semi-supervised} classification. The former is using the few images as \emph{class prototypes}~\cite{snell2017} to estimate class probabilities per region, which are propagated at image level using the \emph{voting process} of WSDDN~\cite{wsddn}. The latter is generating \emph{pseudo-labels} on the unlabeled set from these probabilities to train a classifier~\cite{Lee13}.

By using the PCL pipeline~\cite{tang2018pami} and extending the unlabeled set to both VOC 2007 and VOC 2012, our \ours achieves detection mAP very close to PCL itself trained on VOC 2007 with image-level labels. Moreover, our result is already competitive or superior to many recent WSOD solutions.

It is reasonable to expect further improvement by applying our method to very large unlabeled collections. This is facilitated by the fact that \ours is robust to unknown classes and can discover relevant data even among non-curated collections. Moreover, since \ours produces image-level pseudo-labels that can be used to train any weakly-supervised detection pipeline, further improvement could be expected by using these pseudo-labels with more advanced WSOD methods.

We hope that our work will inspire further research in this challenging regime of limited supervision. A challenge will be to integrate our multi-stage learning process into a single end-to-end trainable pipeline, either including the last WSOD stage (stage 4) or not.

%% file: acknowledgement.tex
\section*{Acknowledgement}

This work was partially supported by the National Natural Science Foundation of China~(NSFC) under Grant No. 61828602 and 61876007.